\documentclass[11pt, a4paper, logo, copyright, nonumbering]{wechat}
\usepackage[numbers, square, sort&compress]{natbib}
\usepackage{dblfloatfix}
\usepackage{ulem}
\usepackage{caption}
\definecolor{citecolor}{HTML}{1976D2}
\hypersetup{
    colorlinks=true,
    linkcolor=black,
    filecolor=magenta,
    urlcolor=blue!50!black,
    citecolor=citecolor,
}
\usepackage{dramatist}
\usepackage{xspace}
\usepackage{pifont}
\usepackage{multirow}
\usepackage{tcolorbox}
\usepackage{xltabular}
\usepackage{longtable}
\usepackage{hyperref}
\usepackage{diagbox}
\usepackage{makecell}
\usepackage{amsmath}   
\usepackage{amssymb}   
\usepackage{amsfonts}  
\usepackage{bm}
\usepackage{lineno}

\usepackage[bottom]{footmisc}

\usepackage{CJKutf8}
\usepackage{subfigure}
\usepackage{setspace}
\usepackage{subcaption} 

\usepackage{titlesec}

\usepackage{tcolorbox}
\usepackage{fontawesome5}
\usepackage{enumitem}

\usepackage{booktabs}
\usepackage{multirow}
\usepackage[table,xcdraw]{xcolor}
\usepackage{graphicx}

\usepackage{algorithm}
\usepackage{algorithmic}
\usepackage{subcaption}
\usepackage{wrapfig}
\definecolor{darkgreen}{RGB}{0, 140, 0}

\tcbset{
    contributionbox/.style={
        colback=blue!3,
        colframe=blue!50!black,
        boxrule=0.5pt,
        arc=3pt,
        left=6pt,
        right=6pt,
        top=6pt,
        bottom=6pt,
        fonttitle=\bfseries
    }
}

\titlespacing*{\paragraph}
  {0pt}                   
  {0.5ex plus 1ex minus .2ex} 
  {1em}            

\makeatletter
\def\@BTrule[#1]{%
  \ifx\longtable\undefined
    \let\@BTswitch\@BTnormal
  \else\ifx\hline\LT@hline
    \nobreak
    \let\@BTswitch\@BLTrule
  \else
     \let\@BTswitch\@BTnormal
  \fi\fi
  \global\@thisrulewidth=#1\relax
  \ifnum\@thisruleclass=\tw@\vskip\@aboverulesep\else
  \ifnum\@lastruleclass=\z@\vskip\@aboverulesep\else
  \ifnum\@lastruleclass=\@ne\vskip\doublerulesep\fi\fi\fi
  \@BTswitch}
\makeatother

\addto\extrasenglish{
}

 {\begin{list}{}%
         {\setlength{\leftmargin}{#1}}%
         \item[]%
 }
 {\end{list}}

\reportnumber{001} 

\title{\centering UniPrefill: Universal Long-Context Prefill Acceleration via Block-wise Dynamic Sparsification}

\author{
    \vspace{0.5em}
    {\large \bfseries
    Qihang Fan$^{1,2,3,\ast}$, Huaibo Huang$^{1, 2,\dagger}$, Zhiying Wu$^{3}$, 
    } \\ \vspace{0.6pt}
    {\large \bfseries
    Bingning Wang$^{3,\ddagger}$, Ran He$^{1, 2}$
    } \\ \vspace{1.0em}
    
    {\normalsize \normalfont
    $^{1}$MAIS\&NLPR, CASIA \quad
    $^{2}$UCAS \quad
    $^{3}$WeChat, Tencent 
    }
}

\renewcommand{\phi}{\varphi}

\renewcommand{\leq}{\leqslant}
\renewcommand{\geq}{\geqslant}

\renewcommand{\epsilon}{\varepsilon}
\renewcommand{\imath}{\mathrm{i}}

\newlength{\restsubwidth}
\newlength{\restsubheight}
\newlength{\restsubmoreheight}
\newcommand{\rest}[2]{%
        \settowidth{\restsubwidth}{\ensuremath{#2}}
        \settoheight{\restsubheight}{\ensuremath{{}_{#2}}}
        \ensuremath{{#1\hskip 0.5pt}_{\vrule\kern2pt\parbox[b][%
        4pt][b]{\the\restsubwidth}{%
                        \ensuremath{{}_{#2}}}}}
        }

\github{https://github.com/qhfan/UniPrefill.git}

\begin{abstract}
As large language models (LLMs) continue to advance rapidly, they are becoming increasingly capable while simultaneously demanding ever-longer context lengths. To improve the inference efficiency of long-context processing, several novel low-complexity hybrid architectures have recently been proposed, effectively alleviating the computational burden of long-context inference. However, existing research on long-context prefill acceleration remains predominantly focused on sparse attention mechanisms, which achieve their maximum speedup only on full-attention models. When transferred to emerging architectures — such as linear/full attention hybrids or sliding window/full attention hybrids — these prefill acceleration approaches suffer significant performance degradation. Furthermore, such methods are generally incompatible with continuous batching, making them difficult to integrate into modern inference engines such as vLLM. To this end, we propose \textbf{UniPrefill}, a prefill acceleration framework applicable to virtually any model architecture, which directly accelerates the model's computation at the token level. We further implement UniPrefill as a continuous batching operator and extend vLLM's scheduling strategy to natively support prefill-decode co-processing and tensor parallel for UniPrefill, enabling its seamless integration into vLLM. UniPrefill achieves up to \textbf{2.1x} speedup in Time-To-First-Token (TTFT), with the acceleration becoming increasingly pronounced as the number of concurrent requests grows.
\end{abstract}

\begin{document}
\begin{CJK*}{UTF8}{gbsn}

\maketitle

\enlargethispage{1cm}

\newcommand\blfootnote[1]{%
  \begingroup
  \renewcommand\thefootnote{}\footnote{#1}%
  \addtocounter{footnote}{-1}%
  \endgroup
}

\blfootnote{$^\dagger$ Corresponding Author.}
\blfootnote{$^\ddagger$ Project Leader.}
\blfootnote{$^\ast$ Work done during internship at WeChat.}

\section{Introduction}

The rapid advancement of large language models (LLMs) has driven their deployment across an increasingly diverse range of real-world applications, from document understanding and code generation to multi-turn dialogue and retrieval-augmented generation~\cite{llama, llama2, qwen2.5-1m, qwen25technicalreport, qwen3technicalreport, qwentechnicalreport, glm2024chatglm}. Alongside this expansion in capability, the context lengths that LLMs are expected to process have grown dramatically — modern deployments routinely involve sequences of tens of thousands of tokens, and the demand for hundred-thousand-token or even million-token contexts is becoming commonplace. This trend places enormous pressure on inference efficiency, as the canonical Softmax Self-Attention~\cite{attention} mechanism scales quadratically with sequence length, incurring prohibitive computational costs when processing long contexts.

To address the quadratic complexity bottleneck, a new generation of hybrid architectures has emerged that interleave computationally efficient layers with full attention layers. Two representative families have gained particular traction: linear/full attention hybrids, which replace a subset of attention layers with linear recurrent mechanisms~\cite{mamba, mamba2, yang2024gla, fan2024rect, fan2024breaking} to reduce per-layer complexity from $O(N^2)$ to $O(N)$; and sliding window/full attention hybrids, which restrict most attention layers to a fixed local context window while retaining a small number of global full-attention layers for long-range dependencies~\cite{gemmateam2025gemma3technicalreport, jiang2023mistral7b}. These hybrid designs substantially reduce the theoretical complexity of long-context inference and have been widely adopted in recently released production-grade models.

\begin{figure}
    \centering
    \includegraphics[width=0.99\linewidth]{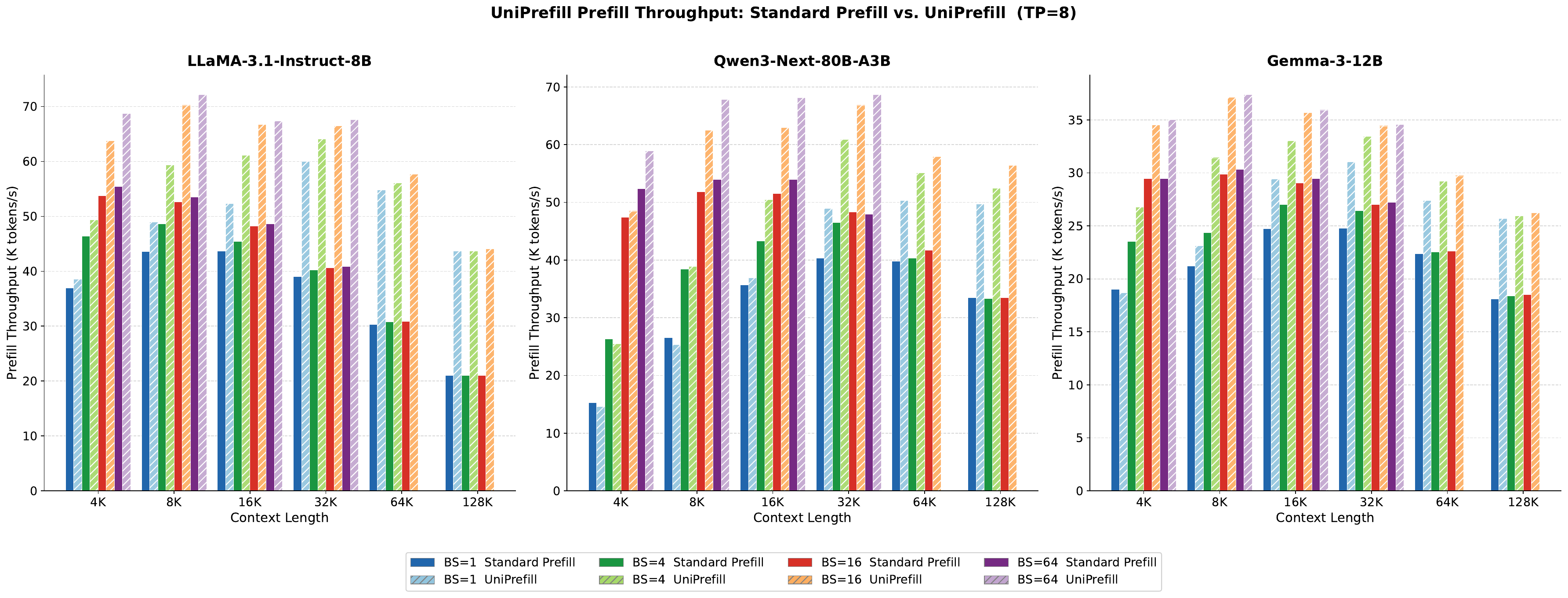}
    \caption{Prefill throughput comparison between Standard Prefill and UniPrefill across three model architectures and varying batch sizes (tensor parallel size is set to 8). All experiments are conducted within vLLM, with UniPrefill deeply integrated into vLLM's continuous batching scheduler. We evaluate prefill throughput (K tokens/s) on LLaMA-3.1-Instruct-8B~\cite{llama3} (full attention), Qwen3-Next-80B-A3B~\cite{qwen3next_blog_2025} (linear/full attention hybrid), and Gemma-3-12B~\cite{gemmateam2025gemma3technicalreport} (sliding window/full attention hybrid) across context lengths from 4K to 128K and batch sizes of 1, 4, 16, and 64. Solid bars denote Standard Prefill and hatched bars denote UniPrefill. UniPrefill consistently achieves higher throughput across all three architectures, with gains becoming more pronounced at longer context lengths and larger batch sizes.}
    \vspace{-4mm}
    \label{fig:uniprefill}
\end{figure}

Despite the proliferation of hybrid architectures, the research community's efforts on prefill acceleration have remained heavily concentrated on sparse attention~\cite{minference, mobamixtureblockattention, fan2026flashprefill}. Representative works such as MInference \cite{minference} have demonstrated impressive prefill speedups, achieving up to 10× acceleration on long sequences under the full-attention-only setting. However, this focus on sparse attention comes with a fundamental limitation: the acceleration is tightly coupled to the full attention operation itself. In hybrid architectures where full attention constitutes only a fraction of all layers, the marginal benefit of accelerating solely those attention layers diminishes considerably. For instance, in a linear/full attention hybrid with a 3:1 ratio, at most one out of every four layers can be accelerated by existing sparse attention methods, leaving the dominant computational budget entirely untouched. This architectural mismatch renders existing prefill acceleration approaches far less effective on the new generation of hybrid models.

A second, equally critical limitation of existing prefill acceleration methods is their incompatibility with continuous batching, the scheduling paradigm that underpins modern high-throughput inference engines such as vLLM \cite{vLLM, zheng2024sglang}. Methods such as FlexPrefill~\cite{flexprefill} operate on individual requests in isolation and assume static batch composition, making them fundamentally difficult to integrate into a continuous batching scheduler where requests enter and exit the batch dynamically. As a result, these methods have largely remained research prototypes and have not been successfully embedded into production inference systems.

To overcome both limitations, we propose UniPrefill, a prefill acceleration framework that achieves architecture-agnostic speedups by exploiting a key insight: token importance can be estimated at full attention layers and propagated across all subsequent layers. Specifically, UniPrefill applies a lightweight block-wise scoring criterion at each full attention layer to identify and drop computationally redundant tokens. Once a token is dropped, it is excluded from all downstream computation in the remaining layers of the block. This cascading effect means that a single token-dropping decision at the attention layer translates into a proportional reduction in computation across the entire layer stack, not merely the attention sublayer. As a result, UniPrefill achieves substantial reductions in both attention FLOPs and GEMM FLOPs simultaneously, making it effective regardless of whether the model is a pure full-attention Transformer or hybrid architecture. 

Beyond the algorithmic design, we address the systems integration challenge by implementing UniPrefill as a continuous batching operator~\cite{yu2022orca} and extending vLLM \cite{vLLM}'s scheduler to natively support prefill-decode co-processing under UniPrefill's token-dropping regime. This tight integration allows UniPrefill to function as a transparent acceleration layer within production inference engines, without requiring changes to model weights or serving infrastructure.

We evaluate UniPrefill on RULER~\cite{hsieh2024ruler} with multiple model architectures. Results demonstrate that UniPrefill introduces no significant accuracy degradation while achieving up to $2.1\times$ speedup in Time-To-First-Token (TTFT), as illustrated in Fig.~\ref{fig:uniprefill}. Notably, the speedup scales favorably with the number of concurrent requests (see Fig.~\ref{fig:uniprefill}), making UniPrefill particularly well-suited for high-concurrency production serving scenarios where prefill cost is the dominant bottleneck.

Our main contributions are summarized as follows:
\begin{itemize}
    \item We propose \textbf{UniPrefill}, a token-level prefill acceleration framework that drops tokens at full attention layers and propagates sparsity across all subsequent layers, reducing both attention and GEMM FLOPs simultaneously, which enables consistent speedups across heterogeneous hybrid architectures.
    \item We implement UniPrefill as a continuous batching operator and integrate it into vLLM \cite{vLLM} via extended scheduling strategies that support prefill-decode co-processing and tensor parallel, enabling seamless production-ready deployment.
    \item Extensive experiments on the long context benchmark RULER demonstrate that UniPrefill achieves up to $2.1\times$ TTFT speedup with negligible accuracy loss, with acceleration gains scaling with request concurrency.
\end{itemize}
\section{Related Works}
\paragraph{Hybrid LLM Architectures.}To overcome the quadratic complexity of Softmax attention, a rich body of work has proposed efficient sequence modeling alternatives, including state space models, linear attention variants, and recurrent architectures~\cite{mamba, mamba2, sun2023retentivenetworksuccessortransformer, yang2024gla, yang2024deltanet, fan2025sec, fan2024rect, minimax01scalingfoundationmodels, yang2024gdn, zhang2025kda}. To balance efficiency and expressiveness, hybrid architectures have emerged that interleave full attention with these efficient alternatives~\cite{qwen3next_blog_2025, lenz2025jamba, gemmateam2025gemma3technicalreport, xiao2026mimov2flash, jiang2023mistral7b}, and have been widely adopted in recently released production models. However, existing prefill acceleration methods remain largely tailored to full-attention-only architectures, limiting their effectiveness on this new generation of models.

\paragraph{Sparse Attention for Prefill Acceleration.}Exploiting the inherent sparsity in attention score matrices is a well-established strategy for accelerating the prefill stage. A body of work identifies static or dynamic sparse patterns — such as vertical, slash, and block-sparse structures — and skips the corresponding attention computations~\cite{minference, native-sparse-attention, mobamixtureblockattention, optimizingmixtureblockattention, flexprefill, chen2026vsprefill}. These methods have demonstrated substantial speedups on full attention models~\cite{minference, flexprefill, xattention, wang2025proxyattn}. However, they share two fundamental limitations: their acceleration is tightly coupled to the attention operation itself, leaving FFN and GEMM computations entirely unaccelerated, and they are generally incompatible with continuous batching~\cite{yu2022orca}, making integration into production inference engines such as vLLM~\cite{vLLM} non-trivial. UniPrefill addresses both limitations by operating at the token level and propagating sparsity across all layers.

\section{Method}

\begin{figure}
    \centering
    \includegraphics[width=0.99\linewidth]{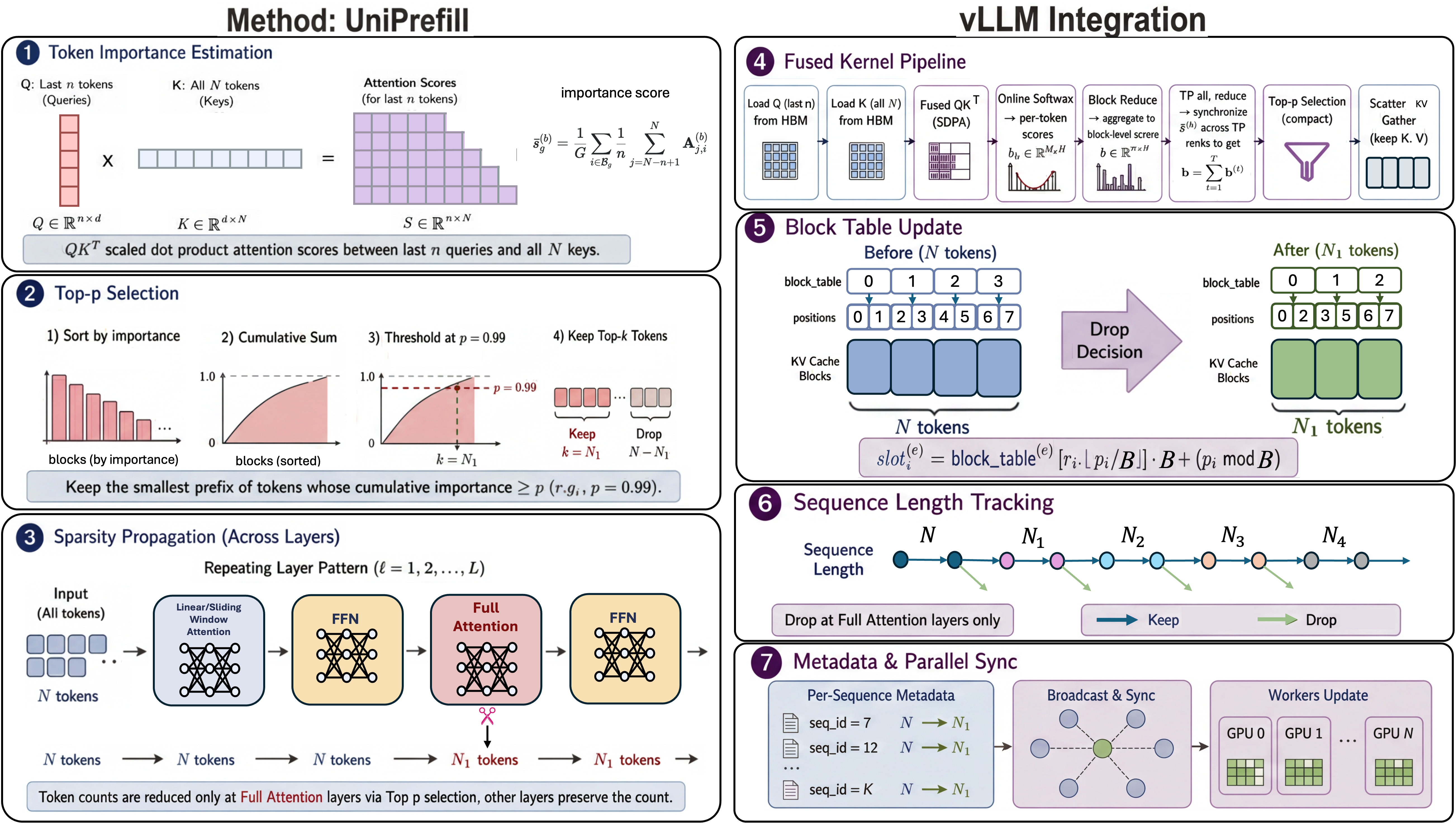}
    \caption{\textbf{Overview of UniPrefill.} \textbf{Left:} UniPrefill estimates token importance via block-level attention scores from the last $n$ queries~\textbf{(1)}, retains the smallest set of token blocks whose cumulative importance reaches $p$~\textbf{(2)}, and propagates the resulting sparsity across all subsequent sub-layers within each repeating layer pattern~\textbf{(3)}. \textbf{Right:} UniPrefill is deeply integrated into vLLM via a fused kernel pipeline~\textbf{(4)}, with KV cache block tables~\textbf{(5)}, per-layer sequence length tracking~\textbf{(6)}, and tensor-parallel metadata synchronisation~\textbf{(7)} updated accordingly.}
    \label{fig:main}
\end{figure}

In this section, we present UniPrefill, an architecture-agnostic
prefill acceleration framework. The overall pipeline is illustrated
in Fig.~\ref{fig:main}.

\subsection{Preliminaries}

Consider an input sequence $\mathbf{x} = [x_1, \ldots, x_N]$ processed by a hybrid LLM consisting of $B$ blocks. Each block $b$ contains a full attention layer followed by $M_b$ sublayers (linear attention, sliding window attention, FFN, etc.). Let $\mathbf{H}^{(b,0)} \in \mathbb{R}^{N \times d}$ denote the block input. The goal of prefill is to compute the final hidden state $\mathbf{h}_N^{(L)}$ for next-token prediction:
\begin{equation}
    P(x_{N+1} \mid x_{1:N}) = \text{LMHead}\!\left(\mathbf{h}_N^{(L)}\right)
\end{equation}
Standard prefill incurs $\mathcal{O}(N^2 d_k)$ per full attention layer and $\mathcal{O}(N d^2)$ per GEMM sublayer, totaling $\mathcal{O}(N^2 d_k + M_b N d^2)$ per block.

\subsection{Token Importance Estimation}

Since next-token prediction depends solely on $\mathbf{h}_N^{(L)}$,
the contribution of token $i$ to the final hidden state at block $b$ is:
\begin{equation}
    \mathbf{h}_N^{(b,1)}
    = \sum_{i=1}^{N} \mathbf{A}^{(b)}_{N,i} \cdot \mathbf{v}_i^{(b)}
    + \mathbf{h}_N^{(b,0)},
\end{equation}
where
$\mathbf{A}^{(b)}_{N,i} =
\operatorname{softmax}_i\!\left(
  \mathbf{q}_N^{(b)} {\mathbf{K}^{(b)}}^\top / \sqrt{d_k}
\right)$
is the full-sequence attention weight.
A token $i$ is negligible to next-token prediction when
$\mathbf{A}^{(b)}_{N,i} \approx 0$.

To reduce estimation variance, we aggregate over the last $n$ query
positions instead of a single position:
\begin{equation}
    s_i^{(b)}
    = \frac{1}{n} \sum_{j=N-n+1}^{N} \mathbf{A}^{(b)}_{j,i},
\end{equation}
requiring an $n \times N$ attention computation at cost
$\mathcal{O}(nNd_k)$, negligible for $n \ll N$.

In practice, importance estimation and token selection operate at
\emph{block granularity}.
We partition the input sequence into non-overlapping blocks of size $G$:
$\mathcal{B}_g = \{(g-1)G+1,\ldots,\min(gG,N)\}$,
$g=1,\ldots,\lceil N/G\rceil$.
For efficiency, the partial GEMM
$\mathbf{S} = \mathbf{Q}_{[N-n:N]}\mathbf{K}^\top \in \mathbb{R}^{n \times N}$
is computed first; an online softmax is then applied \emph{across the
full sequence dimension} to obtain properly normalised attention weights,
after which scores are reduced within each block:
\begin{equation}
    \bar{s}_g^{(b)}
    = \frac{1}{G} \sum_{i \in \mathcal{B}_g}
      \frac{1}{n}\sum_{j=N-n+1}^{N} \mathbf{A}^{(b)}_{j,i},
\end{equation}
where the softmax normalisation is performed over the complete key
sequence before the block reduction, ensuring $\bar{s}_g^{(b)}$
reflects the true attention mass captured by block $g$.
This reduces the number of selection decisions from $N$ to
$\lceil N/G \rceil$ while preserving the accuracy of importance
estimation.

\paragraph{Relationship to SnapKV.}
Our importance estimation shares a surface-level similarity with SnapKV~\cite{li2024snapkv}, which also uses an observation window to identify important tokens. However, the two methods differ fundamentally in objective and scope. SnapKV completes a full $N \times N$ prefill across all layers before applying its selection to compress the KV cache for decode---the prefill FLOPs are entirely unaffected. UniPrefill applies selection \emph{during} prefill, propagating the drop decision forward through all subsequent layers. Formally, whereas SnapKV saves at most $\mathcal{O}(N \cdot d_{kv})$ in decode-time memory per layer, UniPrefill saves $(1-\rho^{(b)}) \cdot M_b \cdot \mathcal{O}(Nd^2)$ in prefill-time FLOPs per block, where $\rho^{(b)} = |\mathcal{S}^{(b)}|/N$ is the token retention ratio---a quantity that grows linearly with $M_b$ and is entirely absent in SnapKV.

\subsection{Top-$p$ Token Selection}

Let $\pi$ be the permutation sorting block-level scores $\{\bar{s}_g^{(b)}\}$ in descending order. We retain the minimal set of blocks:
\begin{equation}
    \mathcal{S}^{(b)} = \left\{\pi(1), \ldots, \pi(k^*)\right\}, \qquad
    k^* = \min k \ \text{ s.t. } \
    \frac{\sum_{j=1}^{k} \bar{s}_{\pi(j)}^{(b)}}{\sum_{g} \bar{s}_g^{(b)}} \geq p
\end{equation}
The dropped set is $\bar{\mathcal{S}}^{(b)} = [N] \setminus \mathcal{S}^{(b)}$. Two structural elements are always retained regardless of their scores: the first $A$ tokens (attention sinks~\cite{xiao2023streamingllm}) and the last $n$ tokens (the query window itself), ensuring causal consistency and numerical stability.

\paragraph{Error bound.}
The perturbation to any retained position $j$ due to dropping $\bar{\mathcal{S}}^{(b)}$ satisfies:
\begin{equation}
    \left\|\Delta \mathbf{h}_j^{(b,1)}\right\|
    \;\leq\;
    \left(\sum_{i \in \bar{\mathcal{S}}^{(b)}} \mathbf{A}^{(b)}_{j,i}\right) \cdot V_{\max}^{(b)}
    \;\leq\;
    (1-p) \cdot V_{\max}^{(b)}
\end{equation}
where $V_{\max}^{(b)} = \max_i \|\mathbf{v}_i^{(b)}\|$. Setting $p = 0.99$ guarantees that at most $1\%$ of the total attention mass is discarded, providing a direct information-theoretic bound on the approximation error at the attention layer.

\paragraph{Top-$p$ vs.\ top-$k$.}
A fixed top-$k$ is insensitive to the actual distribution of attention: when attention is highly concentrated, top-$k$ retains many unnecessary tokens; when diffuse, it may drop tokens with non-trivial contributions. Top-$p$ adapts automatically---the retained set is small when attention is concentrated and large when it is diffuse---providing a uniform bound on approximation error regardless of sequence length or content, which top-$k$ cannot guarantee.

\subsection{Sparsity Propagation Across All Layers}

After token selection at the full attention layer of block $b$, dropped tokens $\bar{\mathcal{S}}^{(b)}$ are excluded from all subsequent sublayers within and beyond the block---every full attention, linear attention, sliding window attention, and FFN layer processes only the retained set $\mathcal{S}^{(b)}$:
\begin{equation}
    \mathbf{H}_{\mathcal{S}}^{(b,m+1)} = f_m\!\left(\mathbf{H}_{\mathcal{S}}^{(b,m)}\right),
    \qquad m = 1, \ldots, M_b
\end{equation}
At block $b+1$, the full sequence is reconstituted by carrying dropped token states forward without update:
\begin{equation}
    \mathbf{H}_i^{(b+1,0)} =
    \begin{cases}
        \mathbf{H}_i^{(b,M_b+1)} & i \in \mathcal{S}^{(b)} \\[4pt]
        \mathbf{H}_i^{(b,0)}     & i \in \bar{\mathcal{S}}^{(b)}
    \end{cases}
\end{equation}
and importance scores are recomputed fresh at each block's full attention layer. This means a single drop decision at layer $\ell$ immediately reduces the token count for all layers $\ell' > \ell$, including subsequent full attention layers, linear attention layers, sliding window layers, and all FFN projections.

\paragraph{FLOPs analysis.}
Let $\mathcal{L}_{\text{drop}} = \{\ell_1, \ell_2, \ldots\}$ denote the set of layers at which dropping is applied, and let $\rho_k$ denote the retention ratio after the $k$-th drop. The total FLOPs saved across all $L$ layers is:
\begin{equation}
    \Delta\mathrm{FLOPs}
    = \sum_{k} (1 - \rho_k) \cdot \sum_{\ell > \ell_k} \mathrm{FLOPs}_\ell(N)
\end{equation}
For a model with $L$ total layers each of cost $\mathcal{O}(Nd^2)$, a single drop at layer $\ell_1$ with retention ratio $\rho$ saves:
\begin{equation}
    \Delta\mathrm{FLOPs}^{(\ell_1)}
    = (1-\rho) \cdot (L - \ell_1) \cdot \mathcal{O}(Nd^2)
\end{equation}
This saving scales linearly with $(L - \ell_1)$, the number of layers remaining after the drop point. Sparse attention methods operating only within the attention sublayer save at most $(1-\rho)\cdot\mathcal{O}(N^2 d_k)$ at that layer alone, leaving all subsequent GEMM costs intact. The ratio of savings is:
\begin{equation}
    \frac{\Delta\mathrm{FLOPs}_{\text{UniPrefill}}}
         {\Delta\mathrm{FLOPs}_{\text{SparseAttn}}}
    = \frac{(L-\ell_1) \cdot Nd^2}{N^2 d_k}
    \xrightarrow{N \to \infty} \infty
\end{equation}
In the long-context regime where $N \gg d$, UniPrefill's GEMM savings dominate, making it particularly effective precisely at the sequence lengths where prefill acceleration matters most.

\paragraph{Error propagation.}
Assuming each sublayer $f_m$ is $L_m$-Lipschitz, the accumulated error at block end satisfies:
\begin{equation}
    \left\|\Delta \mathbf{h}_j^{(b,M_b+1)}\right\|
    \leq (1-p) \cdot V_{\max}^{(b)} \cdot \prod_{m=1}^{M_b} L_m
\end{equation}
Layer normalization and residual connections constrain $\prod_{m} L_m$ in practice, preventing unbounded error amplification across layers.

\subsection{Fused Kernel and vLLM Integration}

\paragraph{Kernel design.}
We implement the importance estimation and top-$p$ selection pipeline as a sequence of four fused kernels operating directly on the variable-length packed token representation indexed by \texttt{cu\_seqlens}, without materializing per-request tensors or padding. The pipeline proceeds as follows:
\begin{equation}
    \mathbf{S} \;=\; \mathbf{Q}_{[N-n:N]} \mathbf{K}^\top \in \mathbb{R}^{n \times N}
    \;\xrightarrow{\text{online softmax}}\;
    \mathbf{o} \in \mathbb{R}^{N}
    \;\xrightarrow{\text{block reduce}}\;
    \mathbf{b} \in \mathbb{R}^{\lceil N/G \rceil}
    \;\xrightarrow{\text{top-}p}\;
    \mathcal{M} \in \{0,1\}^{N}
\end{equation}
The partial GEMM kernel computes $\mathbf{S}$ with tiled $Q$-$K$ blocking and inline causal masking. The softmax kernel aggregates $\operatorname{softmax}(\mathbf{S})$ over the $n$ query rows via a numerically stable two-pass online algorithm, yielding per-token importance scores $\mathbf{o}$. The block-reduce kernel contracts $\mathbf{o}$ across both the head and spatial dimensions within each block of size $G$, producing the block-level score vector $\mathbf{b}$.

The top-$p$ kernel performs sort-and-threshold entirely on-GPU without CPU round-trips. We encode each (score, index) pair into a single \texttt{int64} word via a monotone IEEE-754 bitcast mapping:
\begin{equation}
    \phi(x) =
    \begin{cases}
        \operatorname{bits}(x) \oplus \texttt{0x80000000} & x \geq 0 \\
        \operatorname{bits}(x) \oplus \texttt{0xFFFFFFFF} & x < 0
    \end{cases}
    \quad \Rightarrow \quad
    \texttt{packed} = \bigl(\phi(b_g) \ll 32\bigr) \;\big|\; g
\end{equation}
Sorting packed words descending, computing a cumulative sum of scores, and thresholding at $p$ yields the keep mask $\mathcal{M}$, which is scattered back to original positions. A final expansion kernel lifts $\mathcal{M}$ from block to token granularity, unconditionally setting $\mathcal{M}_i = 1$ for attention-sink tokens $i < A$ and query-window tokens $i \geq N - n$.

\paragraph{Tensor parallelism.}
Under tensor parallelism of degree $T$, each rank observes only $1/T$ of the attention heads, yielding a partial block score $\mathbf{b}^{(t)}$. We synchronize via:
\begin{equation}
    \mathbf{b} = \sum_{t=1}^{T} \mathbf{b}^{(t)}
\end{equation}
before the top-$p$ kernel, ensuring a consistent drop decision across all TP ranks. 

\paragraph{vLLM scheduler integration.}
Integrating token dropping into vLLM's continuous batching scheduler~\cite{yu2022orca,vLLM} requires maintaining correctness across three coupled state structures: layer-wise attention metadata, KV cache slot mappings, and per-request KV length tracking across decode steps.

Upon a drop event at layer $\ell$, we propagate updated metadata to all downstream layers $\ell' > \ell$ by patching \texttt{query\_start\_loc}, \texttt{seq\_lens}, and \texttt{num\_actual\_tokens} to reflect the compacted token stream $|\mathcal{S}^{(\ell)}|$. Physical KV cache slot mappings for each layer $\ell'$ are recomputed as:
\begin{equation}
    \texttt{slot}^{(\ell')}_i
    = \texttt{block\_table}^{(\ell')}\!\left[r_i,\, \lfloor p_i / B \rfloor\right] \cdot B
    + (p_i \bmod B)
\end{equation}
where $p_i$ is the logical position of the $i$-th retained token, $B$ is the KV block size, and $\texttt{block\_table}^{(\ell')}$ is the physical block table of layer $\ell'$---which may differ between global and sliding-window attention layers~\cite{gemmateam2025gemma3technicalreport}.

During decode, each layer $\ell'$ must attend over only the tokens that were physically written to its KV cache during prefill. We maintain a per-request drop history $\{(\ell_k, s_k^r)\}$ recording the retained sequence length $s_k^r$ after each drop event at layer $\ell_k$. The effective KV length visible to layer $\ell'$ during decode is then:
\begin{equation}
    \mathrm{seqused}^{(\ell')}_r
    \;=\; s^{(\ell^-)}_r + \Delta_r, \qquad
    \Delta_r = \mathrm{kv\_len}_r - \mathrm{orig\_len}_r
\end{equation}
where $\ell^- = \max\{\ell_k \in \mathcal{L}_{\mathrm{drop}} : \ell_k < \ell'\}$ is the last drop layer preceding $\ell'$, and $\Delta_r$ counts autoregressive tokens appended since prefill. This per-layer \texttt{seqused} correction is injected into the forward context before each decode step, ensuring every attention layer observes a KV sequence length precisely consistent with its written cache entries---without any modification to model weights or the PagedAttention memory allocator.

\begin{table}[t]
    \centering
    \setlength{\tabcolsep}{0.83mm}
    \scalebox{0.87}{
    \begin{tabular}{c|ccccccc|cccccc}
    \toprule[1pt]
    Method & 4K & 8K & 16K & 32K & 64K & 128K & Avg & 4K & 8K & 16K & 32K & 64K & 128K \\
    \midrule[0.5pt]
    \multicolumn{14}{c}{Llama-3.1-8B-Instruct (Full Attention)} \\
    \midrule[0.5pt]
    Baseline & 97.36 & 95.98 & 94.62 & 91.02 & 86.29 & 76.89 & 90.36 & $1.00\times$ & $1.00\times$ & $1.00\times$ & $1.00\times$ & $1.00\times$ & $1.00\times$\\
    LazyLLM~\cite{lazyllm} & 89.16 & 81.12 & 70.32 & 64.39 & 56.28 & 49.71 & \textcolor{red}{68.50} & $1.09\times$ & $1.19\times$ & $1.28\times$ & $1.74\times$ & $2.21\times$ & $2.51\times$\\
    SlimInfer~\cite{sliminfer} & 90.23 & 82.04 & 71.39 & 67.12 & 57.10 & 45.36 & \textcolor{red}{68.87} & $1.07\times$ & $1.16\times$ & $1.25\times$ & $1.66\times$ & $1.98\times$ & $2.07\times$\\
    MInference~\cite{minference} & 96.71 & 95.78 & 95.51 & 90.76 & 87.12 & 78.21 & 90.68 & $0.82\times$ & $0.86\times$ & $0.98\times$ & $1.03\times$ & $1.15\times$ & $1.34\times$\\
    FlexPrefill~\cite{flexprefill} & 96.34 & 95.12 & 94.83 & 88.96 & 84.31 & 78.13 & 89.62 & $0.83\times$ & $0.89\times$ & $1.02\times$ & $1.08\times$ & $1.24\times$ & $1.46\times$\\
    XAttention~\cite{xattention} & 95.98 & 95.23 & 94.68 & 88.06 & 83.92 & 78.16 & 89.34 & $0.92\times$ & $0.96\times$ & $1.03\times$ & $1.08\times$ & $1.21\times$ & $1.38\times$\\
    ProxyAttn~\cite{wang2025proxyattn} & 96.78 & 95.46 & 95.49 & 89.28 & 85.31 & 78.49 & 90.14 & $0.82\times$ & $0.89\times$ & $1.03\times$ & $1.11\times$ & $1.36\times$ & $1.79\times$\\
    UniPrefill & 96.53 & 95.83 & 95.41 & 89.77 & 85.28 & 79.87 & \textbf{90.45} & $1.21\times$ & $1.34\times$ & $1.37\times$ & $1.62\times$ & $2.01\times$ & $2.26\times$\\
    \midrule[0.5pt]
    \multicolumn{14}{c}{Qwen3-Next-80B-A3B (Linear/Full Attention Hybrid)} \\
    \midrule[0.5pt]
    Baseline & 96.83 & 95.67 & 95.07 & 94.38 & 94.51 & 92.09 & 94.76 & $1.00\times$ & $1.00\times$ & $1.00\times$ & $1.00\times$ & $1.00\times$ & $1.00\times$\\
    LazyLLM~\cite{lazyllm} & 89.13 & 82.37 & 70.69 & 64.37 & 58.12 & 55.17 & \textcolor{red}{69.98} & $1.11\times$ & $1.18\times$ & $1.29\times$ & $1.40\times$ & $1.55\times$ & $1.74\times$\\
    SlimInfer~\cite{sliminfer} & 88.11 & 80.36 & 67.13 & 63.22 & 57.13 & 55.36 & \textcolor{red}{68.55} & $1.14\times$ & $1.22\times$ & $1.27\times$ & $1.34\times$ & $1.42\times$ & $1.56\times$\\
    MInference~\cite{minference} & 96.62 & 94.38 & 94.49 & 94.27 & 94.28 & 91.81 & 94.31 &  $0.96\times$ & $0.98\times$ & $1.00\times$ & $1.00\times$ & $1.02\times$ & $1.05\times$\\
    FlexPrefill~\cite{flexprefill} & 96.36 & 95.03 & 94.17 & 93.91 & 92.89 & 91.44 & 93.97 & $0.96\times$ & $0.98\times$ & $1.00\times$ & $1.01\times$ & $1.04\times$ & $1.08\times$\\
    XAttention~\cite{xattention} & 96.03 & 94.81 & 94.03 & 93.01 & 93.06 & 90.23 & 93.53 & $0.97\times$ & $0.99\times$ & $1.00\times$ & $1.00\times$ & $1.02\times$ & $1.05\times$\\
    ProxyAttn~\cite{wang2025proxyattn} & 96.31 & 94.13 & 94.23 & 93.47 & 93.51 & 91.62 & 93.88 & $0.96\times$ & $0.98\times$ & $1.00\times$ & $1.02\times$ & $1.05\times$ & $1.11\times$\\
    UniPrefill & 96.67 & 94.49 & 94.29 & 93.63 & 93.13 & 91.41 & \textbf{93.94} & $1.08\times$ & $1.21\times$ & $1.24\times$ & $1.39\times$ & $1.42\times$ & $1.68\times$\\
    \midrule[0.5pt]
    \multicolumn{14}{c}{Gemma-3-12B (Sliding Window/Full Attention Hybrid)} \\
    \midrule[0.5pt]
    Baseline & 94.01 & 89.12 & 85.98 & 80.76 & 68.89 & 61.22 & 79.99 & $1.00\times$ & $1.00\times$ & $1.00\times$ & $1.00\times$ & $1.00\times$ & $1.00\times$\\
    LazyLLM~\cite{lazyllm} & 86.11 & 80.34 & 75.23 & 68.42 & 54.12 & 43.38 & \textcolor{red}{67.93} &  $1.23\times$ & $1.32\times$ & $1.37\times$ & $1.42\times$ & $1.53\times$ & $1.64\times$\\
    SlimInfer~\cite{sliminfer} & 88.23 & 83.14 & 79.12 & 69.33 & 53.02 & 40.11 & \textcolor{red}{68.83} & $1.15\times$ & $1.24\times$ & $1.32\times$ & $1.39\times$ & $1.45\times$ & $1.52\times$\\
    MInference~\cite{minference} & 93.56 & 89.51 & 86.01 & 80.04 & 67.09 & 59.31 & 79.25 & $0.98\times$ & $0.99\times$ & $1.00\times$ & $1.00\times$ & $1.01\times$ & $1.03\times$ \\
    FlexPrefill~\cite{flexprefill} & 93.63 & 89.16 & 85.49 & 79.23 & 65.69 & 58.63 & 78.64 & $0.98\times$ & $0.99\times$ & $1.00\times$ & $1.01\times$ & $1.02\times$ & $1.04\times$ \\
    XAttention~\cite{xattention} & 93.06 & 89.24 & 85.67 & 79.14 & 66.18 & 56.24 & 78.26 & $0.99\times$ & $1.00\times$ & $1.01\times$ & $1.01\times$ & $1.02\times$ & $1.02\times$ \\
    ProxyAttn~\cite{wang2025proxyattn} & 93.67 & 89.52 & 85.31 & 78.97 & 65.31 & 59.93 & 78.79 & $0.98\times$ & $0.99\times$ & $1.01\times$ & $1.01\times$ & $1.03\times$ & $1.06\times$ \\
    UniPrefill & 93.18 & 89.76 & 86.47 & 79.08 & 66.32 & 58.38 & \textbf{78.87} & $1.15\times$ & $1.21\times$ & $1.22\times$ & $1.26\times$ & $1.31\times$ & $1.49\times$ \\
    \bottomrule[1pt]
    \end{tabular}}
    \caption{Performance vs. efficiency across different models and methods. Evaluation scores (left) and TTFT speedup relative to baselines (right) are reported. For a fair comparison, all models are evaluated using HuggingFace Transformers with batch size equals to 1.}
    \label{tab:ruler}
\end{table}

\begin{table}[t]
    \centering
    \setlength{\tabcolsep}{1.3mm}
    \scalebox{0.84}{
    \begin{tabular}{c|cccccc|cccccc}
    \toprule[1pt]
    BSZ & 4K & 8K & 16K & 32K & 64K & 128K & 4K & 8K & 16K & 32K & 64K & 128K \\
    \midrule[0.5pt]
    \multicolumn{13}{c}{Llama-3.1-8B-Instruct (Full Attention)} \\
    \midrule[0.5pt]
    1 & 36984 & 43586 & 43697 & 39027 & 30324 & 21013 & \makecell{38522\\(\textcolor{darkgreen}{$+4\%$})} & \makecell{48932\\(\textcolor{darkgreen}{$+12\%$})} & \makecell{52314\\(\textcolor{darkgreen}{$+20\%$})} & \makecell{59984\\(\textcolor{darkgreen}{$+54\%$})} & \makecell{54786\\(\textcolor{darkgreen}{$+81\%$})} & \makecell{43672\\(\textcolor{darkgreen}{$+107\%$})} \\
    4 & 46372 & 48632 & 45436 & 40210 & 30812 & 21054 & \makecell{49336\\(\textcolor{darkgreen}{$+6\%$})} & \makecell{59372\\(\textcolor{darkgreen}{$+22\%$})} & \makecell{61148\\(\textcolor{darkgreen}{$+35\%$})} & \makecell{64113\\(\textcolor{darkgreen}{$+59\%$})} & \makecell{56108\\(\textcolor{darkgreen}{$+82\%$})} & \makecell{43698\\(\textcolor{darkgreen}{$+108\%$})} \\
    16 & 53764 & 52643 & 48221 & 40671 & 30834 & 21062 & \makecell{63762\\(\textcolor{darkgreen}{$+19\%$})} & \makecell{70213\\(\textcolor{darkgreen}{$+33\%$})} & \makecell{66762\\(\textcolor{darkgreen}{$+38\%$})} & \makecell{66512\\(\textcolor{darkgreen}{$+64\%$})} & \makecell{57678\\(\textcolor{darkgreen}{$+87\%$})} & \makecell{44042\\(\textcolor{darkgreen}{$+109\%$})} \\
    64 & 55431 & 53541 & 48603 & 40869 & --- & --- & \makecell{68721\\(\textcolor{darkgreen}{$+24\%$})} & \makecell{72139\\(\textcolor{darkgreen}{$+35\%$})} & \makecell{67361\\(\textcolor{darkgreen}{$+39\%$})} & \makecell{67618\\(\textcolor{darkgreen}{$+65\%$})} & --- & --- \\
    \midrule[0.5pt]
    \multicolumn{13}{c}{Qwen3-Next-80B-A3B (Linear/Full Attention Hybrid)} \\
    \midrule[0.5pt]
    1 & 15314 & 26534 & 35712 & 40312 & 39807 & 33512 & \makecell{14621\\(\textcolor{red}{$-5\%$})} & \makecell{25364\\(\textcolor{red}{$-4\%$})} & \makecell{36912\\(\textcolor{darkgreen}{$+3\%$})} & \makecell{48972\\(\textcolor{darkgreen}{$+21\%$})} & \makecell{50324\\(\textcolor{darkgreen}{$+26\%$})} & \makecell{49732\\(\textcolor{darkgreen}{$+48\%$})} \\
    4 & 26334 & 38442 & 43326 & 46534 & 40322 & 33364 & \makecell{25498\\(\textcolor{red}{$-3\%$})} & \makecell{38872\\(\textcolor{darkgreen}{$+1\%$})} & \makecell{50442\\(\textcolor{darkgreen}{$+16\%$})} & \makecell{60894\\(\textcolor{darkgreen}{$+31\%$})} & \makecell{55136\\(\textcolor{darkgreen}{$+37\%$})} & \makecell{52442\\(\textcolor{darkgreen}{$+57\%$})} \\
    16 & 47432 & 51853 & 51544 & 48303 & 41693 & 33489 & \makecell{48446\\(\textcolor{darkgreen}{$+2\%$})} & \makecell{62468\\(\textcolor{darkgreen}{$+20\%$})} & \makecell{62983\\(\textcolor{darkgreen}{$+22\%$})} & \makecell{66798\\(\textcolor{darkgreen}{$+38\%$})} & \makecell{57938\\(\textcolor{darkgreen}{$+39\%$})} & \makecell{56398\\(\textcolor{darkgreen}{$+68\%$})} \\
    64 & 52334 & 53936 & 53936 & 47932 & --- & --- & \makecell{58936\\(\textcolor{darkgreen}{$+13\%$})} & \makecell{67848\\(\textcolor{darkgreen}{$+26\%$})} & \makecell{68123\\(\textcolor{darkgreen}{$+26\%$})} & \makecell{68631\\(\textcolor{darkgreen}{$+43\%$})} & --- & --- \\
    \midrule[0.5pt]
    \multicolumn{13}{c}{Gemma-3-12B (Sliding Window/Full Attention Hybrid)} \\
    \midrule[0.5pt]
    1 & 19013 & 21203 & 24733 & 24763 & 22367 & 18103 & \makecell{18673\\(\textcolor{red}{$-2\%$})} & \makecell{23132\\(\textcolor{darkgreen}{$+9\%$})} & \makecell{29436\\(\textcolor{darkgreen}{$+19\%$})} & \makecell{31023\\(\textcolor{darkgreen}{$+25\%$})} & \makecell{27384\\(\textcolor{darkgreen}{$+22\%$})} & \makecell{25673\\(\textcolor{darkgreen}{$+42\%$})} \\
    4 & 23531 & 24361 & 27013 & 26432 & 22536 & 18403 & \makecell{26783\\(\textcolor{darkgreen}{$+14\%$})} & \makecell{31432\\(\textcolor{darkgreen}{$+29\%$})} & \makecell{33014\\(\textcolor{darkgreen}{$+22\%$})} & \makecell{33432\\(\textcolor{darkgreen}{$+26\%$})} & \makecell{29232\\(\textcolor{darkgreen}{$+30\%$})} & \makecell{25932\\(\textcolor{darkgreen}{$+41\%$})} \\
    16 & 29468 & 29867 & 29031 & 27012 & 22613 & 18513 & \makecell{34512\\(\textcolor{darkgreen}{$+17\%$})} & \makecell{37136\\(\textcolor{darkgreen}{$+24\%$})} & \makecell{35674\\(\textcolor{darkgreen}{$+23\%$})} & \makecell{34419\\(\textcolor{darkgreen}{$+27\%$})} & \makecell{29732\\(\textcolor{darkgreen}{$+31\%$})} & \makecell{26231\\(\textcolor{darkgreen}{$+42\%$})} \\
    64 & 29471 & 30324 & 29471 & 27236 & --- & --- & \makecell{35016\\(\textcolor{darkgreen}{$+19\%$})} & \makecell{37365\\(\textcolor{darkgreen}{$+23\%$})} & \makecell{35912\\(\textcolor{darkgreen}{$+22\%$})} & \makecell{34578\\(\textcolor{darkgreen}{$+27\%$})} & --- & --- \\
    \bottomrule[1pt]
    \end{tabular}}
    \caption{Prefill throughput (tokens/s) of Standard Prefill and UniPrefill measured within vLLM (TP$=8$) across three model architectures, six context lengths (4K--128K), and four batch sizes (BSZ). The left half of each group reports Standard Prefill throughput; the right half reports UniPrefill throughput.}
    \label{tab:vllminter}
    \vspace{-4mm}
\end{table}

\section{Experiments}
We evaluate UniPrefill across two dimensions: accuracy and efficiency.
For accuracy, we compare UniPrefill against existing prefill acceleration methods on the RULER~\cite{hsieh2024ruler} long-context benchmark across multiple model architectures.
For efficiency, we measure prefill throughput under varying context lengths and batch sizes within our vLLM deployment. Finally, we conduct ablation studies to analyze the contribution of each design choice in UniPrefill. Implementation and deployment details can be found in \textcolor{red}{appendix}.

\subsection{Experimental Setup}
We select three model architectures to validate the effectiveness of UniPrefill: LLaMA-3.1-8B-Instruct~\cite{llama3}, which consists entirely of full-attention layers; Qwen3-Next-80B-A3B~\cite{qwen3next_blog_2025}, a linear/full-attention hybrid with a 3:1 ratio;
and Gemma-3-12B~\cite{gemmateam2025gemma3technicalreport}, a sliding-window/full-attention hybrid with a 5:1 ratio. We set the top-$p$ threshold to $0.99$, $0.99$, and $0.98$ for the three models, respectively.
The minimum dropping granularity is set to a block size of $G=64$ tokens,
and importance scores are estimated using the last $n=128$ query tokens.
To preserve attention sinks~\cite{xiao2023streamingllm}, the first 128 tokens are always retained.

\subsection{Results on RULER}
RULER~\cite{hsieh2024ruler} is a comprehensive long-context benchmark that evaluates LLMs across diverse task categories including retrieval, multi-hop tracing, aggregation, and question answering, with configurable context lengths up to 128K tokens. Unlike prior benchmarks that rely on simple needle-in-a-haystack tests, RULER provides a more rigorous and systematic assessment of true long-context understanding, making it a widely adopted standard for evaluating long-context LLM performance.

Tab.~\ref{tab:ruler} presents RULER scores and TTFT speedups across three model architectures. UniPrefill achieves the best accuracy-efficiency tradeoff among all acceleration methods. LazyLLM and SlimInfer suffer notable accuracy degradation across all three architectures, while sparse attention methods preserve accuracy but yield diminishing speedups on hybrid architectures, with gains often below $1.1\times$ at 128K. UniPrefill strikes the optimal balance: it retains accuracy close to the Baseline while delivering up to $2.26\times$, $1.68\times$, and $1.49\times$ TTFT speedup at 128K context length on LLaMA-3.1-8B, Qwen3-Next-80B-A3B, and Gemma-3-12B, respectively, demonstrating consistent effectiveness across full-attention and hybrid architectures.

\subsection{vLLM Intergration}

Tab.~\ref{tab:vllminter} reports prefill throughput within vLLM across three architectures. UniPrefill consistently improves throughput as context length and batch size increase, achieving up to $+109\%$, $+68\%$, and $+42\%$ gains on LLaMA-3.1-8B, Qwen3-Next-80B-A3B, and Gemma-3-12B, respectively. The speedup scales favorably with both context length and batch size, demonstrating that UniPrefill is particularly effective in the high-concurrency, long-context regime that dominates production serving workloads.

\subsection{Ablation Study}

\paragraph{Block Size.}Tab.~\ref{tab:abblock} presents the ablation results for block size $G \in \{32, 64, 128\}$. At short context lengths, $G=128$ yields the highest speedup due to lower selection overhead per drop decision.
As context length grows, $G=32$ surpasses $G=128$ in speedup, since finer granularity allows more tokens to be dropped, achieving up to $+121\%$ and $+78\%$ throughput gain on LLaMA-3.1-8B and Qwen3-Next-80B-A3B at 128K, respectively. We adopt $G=64$ as the default, which balances selection overhead and drop rate across all context lengths.
\begin{table}[t]
    \centering
    \setlength{\tabcolsep}{0.83mm}
    \scalebox{0.87}{
    \begin{tabular}{c|ccccccc|cccccc}
    \toprule[1pt]
        block size & 4K & 8K & 16K & 32K & 64K & 128K & Avg & 4K & 8K & 16K & 32K & 64K & 128K \\
        \midrule[0.5pt]
        \multicolumn{14}{c}{Llama-3.1-8B-Instruct} \\
        \midrule[0.5pt]
        64 & 96.53 & 95.83 & 95.41 & 89.77 & 85.28 & 79.87 & 90.45 & \textcolor{darkgreen}{$+19\%$} & \textcolor{darkgreen}{$+33\%$} & \textcolor{darkgreen}{$+38\%$} & \textcolor{darkgreen}{$+64\%$} & \textcolor{darkgreen}{$+87\%$} & \textcolor{darkgreen}{$+109\%$}\\
        128 & 94.32 & 93.62 & 93.07 & 88.12 & 83.38 & 78.90 & 88.57 & \textcolor{darkgreen}{$+26\%$} & \textcolor{darkgreen}{$+38\%$} & \textcolor{darkgreen}{$+45\%$} & \textcolor{darkgreen}{$+62\%$} & \textcolor{darkgreen}{$+81\%$} & \textcolor{darkgreen}{$+96\%$}\\
        32 & 93.42 & 94.63 & 95.46 & 90.23 & 85.67 & 79.88 & 89.88 & \textcolor{darkgreen}{$+19\%$} & \textcolor{darkgreen}{$+32\%$} & \textcolor{darkgreen}{$+36\%$} & \textcolor{darkgreen}{$+72\%$} & \textcolor{darkgreen}{$+96\%$} & \textcolor{darkgreen}{$+121\%$}\\
        \midrule[0.5pt]
        \multicolumn{14}{c}{Qwen3-Next-80B-A3B} \\
        \midrule[0.5pt]
        64 & 96.67 & 94.49 & 94.29 & 93.63 & 93.13 & 91.41 & 93.94 & \textcolor{darkgreen}{$+2\%$} & \textcolor{darkgreen}{$+20\%$} & \textcolor{darkgreen}{$+22\%$} & \textcolor{darkgreen}{$+38\%$} & \textcolor{darkgreen}{$+39\%$} & \textcolor{darkgreen}{$+68\%$}\\
        128 & 96.52 & 94.69 & 94.13 & 93.41 & 92.67 & 92.06 & 93.91 & \textcolor{darkgreen}{$+5\%$} & \textcolor{darkgreen}{$+22\%$} & \textcolor{darkgreen}{$+23\%$} & \textcolor{darkgreen}{$+34\%$} & \textcolor{darkgreen}{$+37\%$} & \textcolor{darkgreen}{$+56\%$}\\
        32 & 92.17 & 94.88 & 94.72 & 93.89 & 93.66 & 92.68 & 93.67 & \textcolor{darkgreen}{$+0\%$} & \textcolor{darkgreen}{$+19\%$} & \textcolor{darkgreen}{$+22\%$} & \textcolor{darkgreen}{$+44\%$} & \textcolor{darkgreen}{$+51\%$} & \textcolor{darkgreen}{$+78\%$}\\
        \bottomrule[1pt]
    \end{tabular}}
    \caption{Ablation study of block size $G$. The left panel reports RULER scores under different values of $G$, and the right panel reports the corresponding TTFT speedup of UniPrefill relative to the Baseline.}
    \label{tab:abblock}
\end{table}

\begin{wraptable}{r}{0.5\textwidth}
    \centering
    \vspace{-4mm}
    \setlength{\tabcolsep}{1.2mm}
    \scalebox{0.83}{
    \begin{tabular}{c|ccccccc}
    \toprule[1pt]
        last $n$ & 4K & 8K & 16K & 32K & 64K & 128K & Avg \\
        \midrule[0.5pt]
        \multicolumn{8}{c}{Llama-3.1-8B-Instruct} \\
        \midrule[0.5pt]
        128 & 96.53 & 95.83 & 95.41 & 89.77 & 85.28 & 79.87 & 90.45 \\
        32 & 95.32 & 94.13 & 93.18 & 86.22 & 82.63 & 75.13 & 87.77 \\
        512 & 96.72 & 96.04 & 95.63 & 90.23 & 84.96 & 79.38 & 90.49 \\
    \bottomrule[1pt]
    \end{tabular}}
    \caption{Ablation study of last $n$. RULER scores under different values of the $n$ on LLaMA-3.1-8B-Instruct. $n=128$ is adopted as the default.}
    \vspace{-7mm}
    \label{tab:lastn}
\end{wraptable}

\paragraph{Last $n$.}Tab.~\ref{tab:lastn} reports RULER scores under different values of last $n \in \{32, 128, 512\}$.
$n=32$ leads to a noticeable accuracy drop, as too few query tokens introduce high variance in importance estimation.
$n=512$ recovers accuracy but incurs higher computational overhead.
$n=128$ achieves the best balance and is adopted as the default.

\section{Conclusion}
We present UniPrefill, an architecture-agnostic framework for long-context LLM prefill acceleration. By estimating token importance via block-wise top-$p$ selection at full-attention layers and propagating the sparsity mask across all subsequent sub-layers, UniPrefill simultaneously reduces attention and GEMM FLOPs, making it effective across full-attention and hybrid architectures alike. We implement UniPrefill as a fused kernel pipeline and integrate it into vLLM's continuous-batching scheduler without any model weight changes. Experiments on the RULER benchmark show that UniPrefill achieves the best accuracy-efficiency tradeoff among all compared methods, delivering up to $2.1\times$ TTFT speedup with negligible accuracy loss, with gains scaling favorably with context length and batch size. We hope UniPrefill provides a practical and general solution for efficient long-context LLM serving.

\bibliography{main}

\newpage
\appendix

\appendix
\section{Implementation and deployment details}

UniPrefill is implemented and evaluated on top of vLLM v0.16.0~\cite{vLLM}, with full support for prefill-decode co-processing and tensor parallelism, making it compatible with standard production deployment configurations. All throughput experiments are conducted under tensor parallelism degree $\text{TP}=8$, reflecting a typical multi-GPU serving setup in real-world scenarios. The continuous-batching operator is implemented as a set of fused Triton kernels, which are hardware-agnostic by design and theoretically portable across different accelerator platforms. All experiments are conducted under CUDA 12.8.

\section{Experiment statistical significance.}
Tab.~\ref{tab:abrs} reports results across multiple random seeds, and the consistently stable performance demonstrates that UniPrefill is robust to random seed initialization.
\begin{table}[h]
    \centering
    \begin{tabular}{c|cccccc}
    \toprule[1pt]
        random seed & 4K & 8K & 16K & 32K & 64K & 128K \\
        \midrule[0.5pt]
        0 & 96.53 & 95.83 & 95.41 & 89.77 & 85.28 & 79.87 \\
        321 & 96.53 & 95.83 & 95.41 & 89.77 & 85.28 & 79.87 \\
        3467 & 96.53 & 95.83 & 95.41 & 89.77 & 85.28 & 79.87 \\
    \bottomrule[1pt]
    \end{tabular}
    \caption{Ablation study on different random seed.}
    \label{tab:abrs}
\end{table}

\section{Limitations and Broader Impacts}
This work focuses on accelerating the prefill phase for long-context LLM inference. While UniPrefill delivers consistent speedups across diverse architectures and sequence lengths without observable accuracy degradation, extending the framework to broader inference optimization dimensions—such as decoding acceleration or training-time efficiency—remains an interesting direction for future work. Broader societal considerations, including ethical deployment and safety alignment, are important but lie outside the technical scope of this study.

\end{CJK*}
\end{document}